# Neural Aggregation Network for Video Face Recognition

Jiaolong Yang[*,1,2,3], Peiran Ren[1], Dongqing Zhang[1], Dong Chen[1], Fang Wen[1], Hongdong Li[2], Gang Hua[1]

[1]Microsoft Research   [2]The Australian National University   [3]Beijing Institute of Technology

## Abstract

*This paper presents a Neural Aggregation Network (NAN) for video face recognition. The network takes a face video or face image set of a person with a variable number of face images as its input, and produces a compact, fixed-dimension feature representation for recognition. The whole network is composed of two modules. The feature embedding module is a deep Convolutional Neural Network (CNN) which maps each face image to a feature vector. The aggregation module consists of two attention blocks which adaptively aggregate the feature vectors to form a single feature inside the convex hull spanned by them. Due to the attention mechanism, the aggregation is invariant to the image order. Our NAN is trained with a standard classification or verification loss without any extra supervision signal, and we found that it automatically learns to advocate high-quality face images while repelling low-quality ones such as blurred, occluded and improperly exposed faces. The experiments on IJB-A, YouTube Face, Celebrity-1000 video face recognition benchmarks show that it consistently outperforms naive aggregation methods and achieves the state-of-the-art accuracy.*

## 1. Introduction

Video face recognition has caught more and more attention from the community in recent years [42, 21, 43, 11, 26, 22, 23, 27, 15, 35, 31, 10]. Compared to image-based face recognition, more information of the subjects can be exploited from the input videos, which naturally incorporate faces of the same subject in varying poses and illumination conditions. The key issue in video face recognition is to build an appropriate representation of the video face, such that it can effectively integrate the information across different frames together, maintaining beneficial while discarding noisy information.

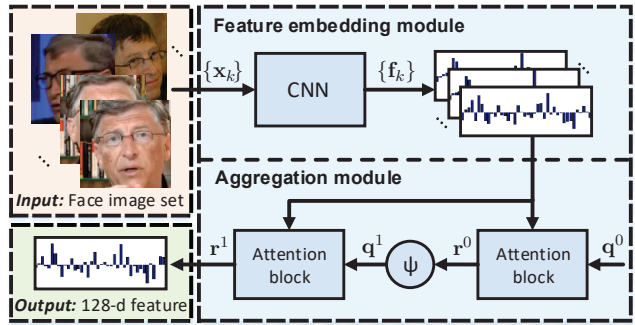

Figure 1. Our network architecture for video face recognition. All input face images $\{\mathbf{x}_k\}$ are processed by a feature embedding module with a deep CNN, yielding a set of feature vectors $\{\mathbf{f}_k\}$. These features are passed to the aggregation module, producing a single 128-dimensional vector $\mathbf{r}^1$ to represent the input faces images. This compact representation is used for recognition.

One naive approach would be representing a video face as a set of frame-level face features such as those extracted from deep neural networks [35, 31], which have dominated face recognition recently [35, 28, 33, 31, 24, 41]. Such a representation comprehensively maintains the information across all frames. However, to compare two video faces, one needs to fuse the matching results across all pairs of frames between the two face videos. Let $n$ be the average number of video frames, then the computational complexity is $O(n^2)$ per match operation, which is not desirable especially for large-scale recognition. Besides, such a set-based representation would incur $O(n)$ space complexity per video face example, which demands a lot of memory storage and confronts efficient indexing.

We argue that it is more desirable to come with a compact, *fixed-size* feature representation at the video level, irrespective of the varied length of the videos. Such a representation would allow direct, *constant-time* computation of the similarity or distance without the need for frame-to-frame matching. A straightforward solution might be extracting a feature at each frame and then conducting a certain type of pooling to aggregate the frame-level features together to form a video-level representation.

---

The most commonly adopted pooling strategies may be average and max pooling [28, 22, 7, 9]. While these naive pooling strategies were shown to be effective in the previous works, we believe that a good pooling or aggregation strategy should *adaptively* weigh and combine the frame-level features across all frames. The intuition is simple: a video (especially a long video sequence) or an image set may contain face images captured at various conditions of lighting, resolution, head pose *etc.*, and a smart algorithm should favor face images that are more discriminative (or more "memorizable") and prevent poor face images from jeopardizing the recognition.

To this end, we look for an adaptive weighting scheme to linearly combine all frame-level features from a video together to form a compact and discriminative face representation. Different from the previous methods, we neither fix the weights nor rely on any particular heuristics to set them. Instead, we designed a neural network to adaptively calculate the weights. We named our network the Neural Aggregation Network (NAN), whose coefficients can be trained through supervised learning in a normal face recognition training task without the need for extra supervision signals.

The proposed NAN is composed of two major modules that could be trained end-to-end or one by one separately. The first one is a feature embedding module which serves as a frame-level feature extractor using a deep CNN model. The other is the aggregation module that adaptively fuses the feature vectors of all the video frames together.

Our neural aggregation network is designed to inherit the main advantages of pooling techniques, including the ability to handle arbitrary input size and producing order-invariant representations. The key component of this network is inspired by the Neural Turing Machine [12] and the work of [38], both of which applied an attention mechanism to organize the input through accessing an external memory. This mechanism can take an input of arbitrary size and work as a tailor emphasizing or suppressing each input element just via a weighted averaging, and very importantly it is order independent and has trainable parameters. In this work, we design a simple network structure of two cascaded attention blocks associated with this attention mechanism for face feature aggregation.

Apart from building a video-level representation, the neural aggregation network can also serve as a subject level feature extractor to fuse multiple data sources. For example, one can feed it with all available images and videos, or the aggregated video-level features of multiple videos from the same subject, to obtain a single feature representation with fixed size. In this way, the face recognition system not only enjoys the time and memory efficiency due to the compact representation, but also exhibits superior performance, as we will show in our experiments.

We evaluated the proposed NAN for both the tasks of video face verification and identification. We observed consistent margins in three challenging datasets, including the YouTube Face dataset [42], the IJB-A dataset [19], and the Celebrity-1000 dataset [23], compared to the baseline strategies and other competing methods.

Last but not least, we shall point out that our proposed NAN can serve as a general framework for learning content-adaptive pooling. Therefore, it may also serve as a feature aggregation scheme for other computer vision tasks.

### 1.1. Related Works

Face recognition based on videos or image sets has been actively studied in the past. This paper is concerned with the input being an orderless set of face images. Existing methods exploiting temporal dynamics will not be considered here. For set based face recognition, many previous methods have attempted to represent the set of face images with appearance subspaces or manifolds and perform recognition via computing manifold similarity or distance [20, 2, 18, 40, 37]. These traditional methods may work well under constrained settings but usually cannot handle the challenging unconstrained scenarios where large appearance variations are present.

Along a different axis, some methods build video feature representation based on local features [21, 22, 27]. For example, the PEP methods [21, 22] take a part-based representation by extracting and clustering local features. The Video Fisher Vector Faces ($VF^2$) descriptor [27] uses Fisher Vector coding to aggregate local features across different video frames together to form a video-level representation.

Recently, state-of-the-art face recognition methods has been dominated by deep convolution neural networks [35, 31, 28, 7, 9]. For video face recognition, most of these methods either use pairwise frame feature similarity computation [35, 31] or naive (average/max) frame feature pooling [28, 7, 9]. This motivated us to seek for an adaptive aggregation approach.

As previously mentioned, this work is also related to the Neural Turing Machine [12] and the work of [38]. However, it is worth noting that while they use Recurrent Neural Networks (RNN) to handle sequential inputs/outputs, there is *no* RNN structure in our method. We only borrow their differentiable memory addressing/attention scheme for our feature aggregation.

## 2. Neural Aggregation Network

As shown in Fig. 1, the NAN network takes a set of face images of a person as input and outputs a single feature vector as its representation for the recognition task. It is built upon a modern deep CNN model for frame feature embedding, and becomes more powerful for video face recognition by adaptively aggregating all frames in the video into a compact vector representation.

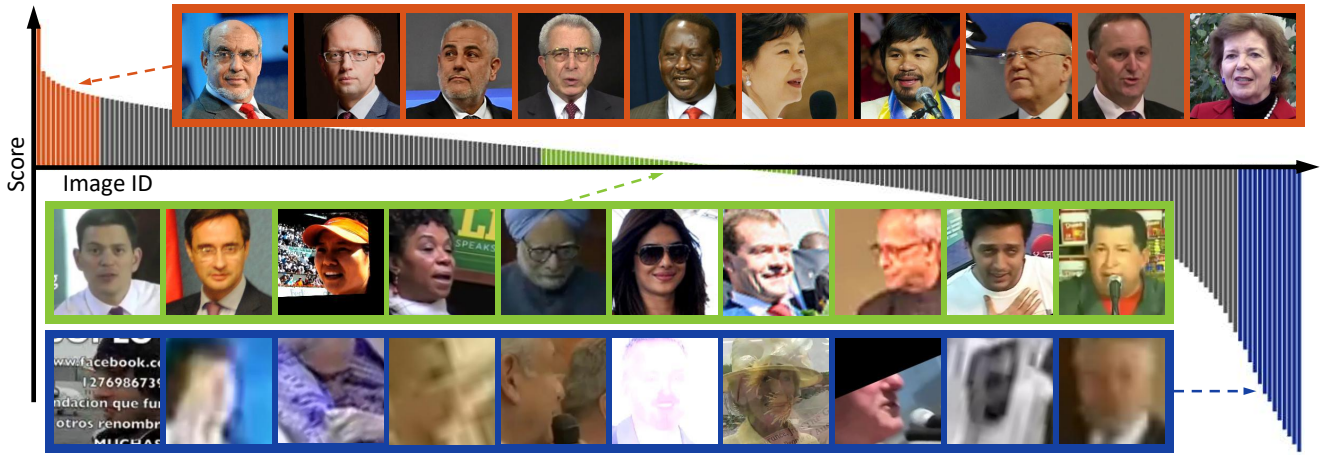

Figure 2. Face images in the IJB-A dataset, sorted by their scores (values of $e$ in Eq. 2) from a *single attention block* trained in the face recognition task. The faces in the top, middle and bottom rows are sampled from the faces with scores in the highest 5%, a 10% window centered at the median, and the lowest 5%, respectively.

### 2.1. Feature embedding module

The image embedding module of our NAN is a deep Convolution Neural Network (CNN), which embeds each frame of a video to a face feature representation. To leverage modern deep CNN networks with high-end performances, in this paper we adopt the GoogLeNet [34] with the Batch Normalization (BN) technique [17]. Certainly, other network architectures are equally applicable here as well. The GoogLeNet produces 128-dimension image features, which are first *normalized* to be unit vectors then fed into the aggregation module. In the rest of this paper, we will simply refer to the employed GoogLeNet-BN network as CNN.

### 2.2. Aggregation module

Consider the video face recognition task on $n$ pairs of video face data $(\mathcal{X}^i, y_i)_{i=1}^n$, where $\mathcal{X}^i$ is a face video sequence or a image set with varying image number $K_i$, *i.e.* $\mathcal{X}^i = \{\mathbf{x}_1^i, \mathbf{x}_2^i, ..., \mathbf{x}_{K_i}^i\}$ in which $\mathbf{x}_k^i, k = 1, ..., K_i$ is the $k$-th frame in the video, and $y_i$ is the corresponding subject ID of $\mathcal{X}^i$. Each frame $\mathbf{x}_k^i$ has a corresponding normalized feature representation $\mathbf{f}_k^i$ extracted from the feature embedding module. For better readability, we omit the upper index where appropriate in the remaining text. Our goal is to utilize all feature vectors from a video to generate a set of linear weights $\{a_k\}_{k=1}^K$, so that the aggregated feature representation becomes

$$\mathbf{r} = \sum_k a_k \mathbf{f}_k. \tag{1}$$

In this way, the aggregated feature vector has the same size as a single face image feature extracted by the CNN.

Obviously, the key of Eq. 1 is its weights $\{a_k\}$. If $a_k \equiv \frac{1}{K}$, Eq. 1 will degrades to naive averaging, which is usually non-optimal as we will show in our experiments. We instead try to design a better weighting scheme.

Three main principles have been considered in designing our aggregation module. First, the module should be able to process different numbers of images (*i.e.* different $K_i$'s), as the video data source varies from person to person. Second, the aggregation should be invariant to the image order – we prefer the result unchanged when the image sequence are reversed or reshuffled. This way, the aggregation module can handle an arbitrary set of image or video faces without temporal information (*e.g.* that collected from different Internet locations). Third, the module should be adaptive to the input faces and has parameters trainable through supervised learning in a standard face recognition training task.

Our solution is inspired by the *memory attention mechanism* described in [12, 32, 38]. The idea therein is to use a neural model to read external memories through a differentiable addressing/attention scheme. Such models are often coupled with Recurrent Neural Networks (RNN) to handle sequential inputs/outputs [12, 32, 38]. Although an RNN structure is not needed for our purpose, its memory attention mechanism is applicable to our aggregation task. In this work, we treat the face features as the memory and cast feature weighting as a memory addressing procedure. We employ in the aggregation module the "attention blocks", to be described as follows.

#### 2.2.1 Attention blocks

An attention block reads all feature vectors from the feature embedding module, and generate linear weights for them. Specifically, let $\{\mathbf{f}_k\}$ be the face feature vectors, then an attention block filters them with a kernel $\mathbf{q}$ via dot product, yielding a set of corresponding significances $\{e_k\}$. They are then passed to a softmax operator to generate positive weights $\{a_k\}$ with $\sum_k a_k = 1$. These two operations can be described by the following equations, respectively:

$$e_k = \mathbf{q}^T \mathbf{f}_k \quad (2)$$

$$a_k = \frac{\exp(e_k)}{\sum_j \exp(e_j)}. \quad (3)$$

Table 1. Performance comparison on the IJB-A dataset. TAR/FAR: True/False Accept Rate for verification. TPIR/FPIR: True/False Positive Identification Rate for identification.

| Method | 1:1 Verification TAR@FAR of: | | 1:N Identification TPIR@FPIR of: | |
|---|---|---|---|---|
| | 0.001 | 0.01 | 0.01 | 0.1 |
| CNN+AvgPool | 0.771 | 0.913 | 0.634 | 0.879 |
| NAN single attention | 0.847 | 0.927 | 0.778 | 0.902 |
| NAN cascaded attention | 0.860 | 0.933 | 0.804 | 0.909 |

It can be seen that our algorithm essentially selects one point inside of the convex hull spanned by all the feature vectors. One related work is [3] where each face image set is approximated with a convex hull and set similarities are defined as the shortest path between two convex hulls.

In this way, the number of inputs $\{\mathbf{f}_k\}$ does not affect the size of aggregation $\mathbf{r}$, which is of the same dimension as a single feature $\mathbf{f}_k$. Besides, the aggregation result is invariant to the input order of $\mathbf{f}_k$: according to Eq. 1, 2, and 3, permuting $\mathbf{f}_k$ and $\mathbf{f}_{k'}$ has no effects on the aggregated representation $\mathbf{r}$. Furthermore, an attention block is modulated by the filter kernel $\mathbf{q}$, which is trainable through standard backpropagation and gradient descent.

**Single attention block – Universal face feature quality measurement.** We first try using one attention block for aggregation. In this case, vector $\mathbf{q}$ is the parameter to learn. It has the same size as a single feature $\mathbf{f}$ and serves as a universal prior measuring the face feature quality.

We train the network to perform video face verification (see Section 2.3 and Section 3 for details) in the IJB-A dataset [19] on the extracted face features, and Figure 2 shows the sorted scores of all the faces images in the dataset. It can be seen that after training, the network favors high-quality face images, such as those of high resolutions and with relatively simple backgrounds. It down-weights face images with blur, occlusion, improper exposure and extreme poses. Table 1 shows that the network achieves higher accuracy than the average pooling baseline in the verification and identification tasks.

**Cascaded two attention blocks – Content-aware aggregation.** We believe a content-aware aggregation can perform even better. The intuition behind is that face image variation may be expressed differently at different geo-graphic locations in the feature space (*i.e.* for different persons), and content-aware aggregation can learn to select features that are more discriminative for the identity of the input image set. To this end, we employ two attention blocks in a cascaded and end-to-end fashion described as follows.

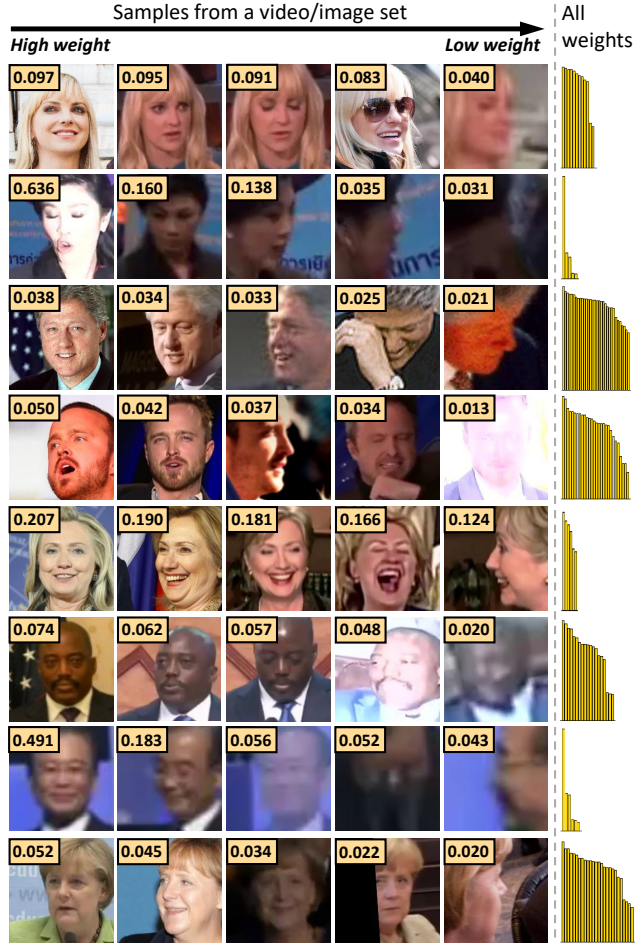

Figure 3. Typical examples showing the weights of the images in the image sets computed by our NAN. In each row, five faces images are sampled from an image set and sorted based on their weights (numbers in the rectangles); the rightmost bar chart shows the sorted weights of all the images in the set (heights scaled).

Let $\mathbf{q}^0$ be the kernel of the first attention block, and $\mathbf{r}^0$ be the aggregated feature with $\mathbf{q}^0$. We adaptively compute $\mathbf{q}^1$, the kernel of the second attention block, through a transfer layer taking $\mathbf{r}^0$ as the input:

$$\mathbf{q}^1 = \tanh(\mathbf{W}\mathbf{r}^0 + \mathbf{b}) \quad (4)$$

where $\mathbf{W}$ and $\mathbf{b}$ are the weight matrix and bias vector of the neurons respectively, and $\tanh(x) = \frac{e^x - e^{-x}}{e^x + e^{-x}}$ imposes the hyperbolic tangent nonlinearity. The feature vector $\mathbf{r}^1$ generated by $\mathbf{q}^1$ will be the final aggregation results. Therefore, $(\mathbf{q}^0, \mathbf{W}, \mathbf{b})$ are now the trainable parameters of the aggregation module.

We train the network on the IJB-A dataset again, and Table 1 shows that the network obtained better results than using single attention block. Figure 3 shows some typical examples of the weights computed by the trained network for different videos or image sets.

Our current full solution of NAN, based on which all the remaining experimental results are obtained, adopts such a cascaded two attention block design (as per Fig. 1).

### 2.3. Network training

The NAN network can be trained either for face verification and identification tasks with standard configurations.

#### 2.3.1 Training loss

For verification, we build a *siamese* neural aggregation network structure [8] with two NANs sharing weights, and minimize the average contrastive loss [14]: $l_{i,j} = y_{i,j}||\mathbf{r}_i^1 - \mathbf{r}_j^1||_2^2 + (1-y_{i,j})\max(0, m - ||\mathbf{r}_i^1 - \mathbf{r}_j^1||_2^2)$, where $y_{i,j} = 1$ if the pair $(i, j)$ is from the same identity and $y_{i,j} = 0$ otherwise. The constant $m$ is set to 2 in all our experiments.

For identification, we add on top of NAN a fully-connected layer followed by a softmax and minimize the average classification loss.

#### 2.3.2 Module training

The two modules can be trained either simultaneously in an end-to-end fashion, or separately one by one. The latter option is chosen in this work. Specifically, we first train the CNN on single images with the identification task, then we train the aggregation module on top of the features extracted by CNN. More details can be found in Section 3.1.

We chose this separate training strategy mainly for two reasons. First, in this work we would like to focus on analyzing the effectiveness and performance of the aggregation module with the attention mechanism. Despite the huge success of applying deep CNN in image-based face recognition task, little attention has been drawn to CNN feature aggregation to our knowledge. Second, training a deep CNN usually necessitates a large volume of labeled data. While millions of still images can be obtained for training nowadays [35, 28, 31], it appears not practical to collect such amount of distinctive face videos or sets. We leave an end-to-end training of the NAN as our future work.

## 3. Experiments

This section evaluates the performance of the proposed NAN network. We will begin with introducing our training details and the baseline methods, followed by reporting the results on three video face recognition datasets: the IARPA Janus Benchmark A (IJB-A) [19], the YouTube Face dataset [42], and the Celebrity-1000 dataset [23].

### 3.1. Training details

As mentioned in Section 2.3, two networks are trained separately in this work. To train the CNN, we use about 3M face images of 50K identities crawled from the Internet to perform image-based identification. The faces are detected using the JDA method [5], and aligned with the LBF method [29]. The input image size is 224x224. After training, the CNN is fixed and we focus on analyzing the effectiveness of the neural aggregation module.

The aggregation module is trained on each video face dataset we tested on with standard backpropagation and an RMSProp solver [36]. An all-zero parameter initialization is used, *i.e.*, we start from average pooling. The batch size, learning rate, and iteration are tuned for each dataset. As the network is quite simple and image features are compact (128-d), the training process is quite efficient: training on 5K video pairs with ∼1M images in total only takes less than 2 minutes on a CPU of a desktop PC.

### 3.2. Baseline methods

Since our goal is compact video face representation, we compare the results with simple aggregation strategies such as average pooling. We also compare with some set-to-set similarity measurements leveraging pairwise comparison on the image level. To keep it simple, *we simply use the $L_2$ feature distances for face recognition* (all features are normalized), although it is possible to combine with an extra metric learning or template adaption technique [10] to further boost the performance on each dataset.

In the baseline methods, *CNN+Min $L_2$*, *CNN+Max $L_2$*, *CNN+Mean $L_2$* and *CNN+SoftMin $L_2$* measure the similarity of two video faces based on the $L_2$ feature distances of all frame pairs. They necessitate storing all image features of a video, *i.e.*, with $O(n)$ space complexity. The first three use the minimum, maximum and mean pairwise distance respectively, thus having $O(n^2)$ complexity for similarity computation. *CNN+SoftMin $L_2$* corresponds to the SoftMax similarity score advocated in some works such as [24, 25, 1]. It has $O(m \cdot n^2)$ complexity for computation[1].

*CNN+MaxPool* and *CNN+AvePool* are respectively max-pooling and average-pooling along each feature dimension for aggregation. These two methods as well as our NAN produce a 128-d feature representation for each video and compute the similarity in $O(1)$ time.

### 3.3. Results on IJB-A dataset

The IJB-A dataset [19] contains face images and videos captured from unconstrained environments. It features full pose variation and wide variations in imaging conditions thus is very challenging. There are 500 subjects with 5,397 images and 2,042 videos in total and 11.4 images and 4.2 videos per subject on average. We detect the faces with landmarks using STN [4] face detector, and then align the face image with similarity transformation.

---
[1]$m$ is the number of scaling factor $\beta$ used (see [25] for details). We tested 20 combinations of (negative) $\beta$'s, including single [1] or multiple values [24, 25] and report the best results obtained.

Table 2. Performance evaluation on the IJB-A dataset. For verification, the true accept rates (TAR) *vs.* false positive rates (FAR) are reported. For identification, the true positive identification rate (TPIR) *vs.* false positive identification rate (FPIR) and the Rank-N accuracies are presented. ($\dagger$: first aggregating the images in each media then aggregate the media features in a template. $*$: results cited from [10].)

| Method | 1:1 Verification TAR | | | 1:N Identification TPIR | | | | |
|---|---|---|---|---|---|---|---|---|
| | FAR=0.001 | FAR=0.01 | FAR=0.1 | FPIR=0.01 | FPIR=0.1 | Rank-1 | Rank-5 | Rank-10 |
| B-CNN [9] | – | – | – | $0.143 \pm 0.027$ | $0.341 \pm 0.032$ | $0.588 \pm 0.020$ | $0.796 \pm 0.017$ | – |
| LSFS [39] | $0.514 \pm 0.060$ | $0.733 \pm 0.034$ | $0.895 \pm 0.013$ | $0.383 \pm 0.063$ | $0.613 \pm 0.032$ | $0.820 \pm 0.024$ | $0.929 \pm 0.013$ | – |
| $DCNN_{manual}$+metric[7] | – | $0.787 \pm 0.043$ | $0.947 \pm 0.011$ | – | – | $0.852 \pm 0.018$ | $0.937 \pm 0.010$ | $0.954 \pm 0.007$ |
| Triplet Similarity [30] | $0.590 \pm 0.050$ | $0.790 \pm 0.030$ | $0.945 \pm 0.002$ | $0.556 \pm 0.065*$ | $0.754 \pm 0.014*$ | $0.880 \pm 0.015*$ | $0.95 \pm 0.007$ | $0.974 \pm 0.005*$ |
| Pose-Aware Models [24] | $0.652 \pm 0.037$ | $0.826 \pm 0.018$ | – | – | – | $0.840 \pm 0.012$ | $0.925 \pm 0.008$ | $0.946 \pm 0.007$ |
| Deep Milti-Pose [1] | – | 0.876 | 0.954 | $0.52*$ | $0.75*$ | 0.846 | 0.927 | 0.947 |
| $DCNN_{fusion}$ [6] | – | $0.838 \pm 0.042$ | $0.967 \pm 0.009$ | $0.577 \pm 0.094*$ | $0.790 \pm 0.033*$ | $0.903 \pm 0.012$ | $0.965 \pm 0.008$ | $0.977 \pm 0.007$ |
| Masi *et al.* [25] | 0.725 | 0.886 | – | – | – | 0.906 | 0.962 | 0.977 |
| Triplet Embedding [30] | $0.813 \pm 0.02$ | $0.90 \pm 0.01$ | $0.964 \pm 0.005$ | $0.753 \pm 0.03$ | $0.863 \pm 0.014$ | $0.932 \pm 0.01$ | – | $0.977 \pm 0.005$ |
| VGG-Face [28] | – | $0.805 \pm 0.030*$ | – | $0.461 \pm 0.077*$ | $0.670 \pm 0.031*$ | $0.913 \pm 0.011*$ | – | $0.981 \pm 0.005*$ |
| Template Adaptation[10] | $0.836 \pm 0.027$ | $0.939 \pm 0.013$ | **$0.979 \pm 0.004$** | $0.774 \pm 0.049$ | $0.882 \pm 0.016$ | $0.928 \pm 0.010$ | $0.977 \pm 0.004$ | **$0.986 \pm 0.003$** |
| CNN+Max $L_2$ | $0.202 \pm 0.029$ | $0.345 \pm 0.025$ | $0.601 \pm 0.024$ | $0.149 \pm 0.033$ | $0.258 \pm 0.026$ | $0.429 \pm 0.026$ | $0.632 \pm 0.033$ | $0.722 \pm 0.030$ |
| CNN+Min $L_2$ | $0.038 \pm 0.008$ | $0.144 \pm 0.073$ | $0.972 \pm 0.006$ | $0.026 \pm 0.009$ | $0.293 \pm 0.175$ | $0.853 \pm 0.012$ | $0.903 \pm 0.010$ | $0.924 \pm 0.009$ |
| CNN+Mean $L_2$ | $0.688 \pm 0.080$ | $0.895 \pm 0.016$ | $0.978 \pm 0.004$ | $0.514 \pm 0.116$ | $0.821 \pm 0.040$ | $0.916 \pm 0.012$ | $0.973 \pm 0.005$ | $0.980 \pm 0.004$ |
| CNN+SoftMin $L_2$ | $0.697 \pm 0.085$ | $0.904 \pm 0.015$ | $0.978 \pm 0.004$ | $0.500 \pm 0.134$ | $0.831 \pm 0.039$ | $0.919 \pm 0.010$ | $0.973 \pm 0.005$ | $0.981 \pm 0.004$ |
| CNN+MaxPool | $0.202 \pm 0.029$ | $0.345 \pm 0.025$ | $0.601 \pm 0.024$ | $0.079 \pm 0.005$ | $0.179 \pm 0.020$ | $0.757 \pm 0.025$ | $0.911 \pm 0.013$ | $0.945 \pm 0.009$ |
| CNN+AvePool | $0.771 \pm 0.064$ | $0.913 \pm 0.014$ | $0.977 \pm 0.004$ | $0.634 \pm 0.109$ | $0.879 \pm 0.023$ | $0.931 \pm 0.011$ | $0.972 \pm 0.005$ | $0.979 \pm 0.004$ |
| CNN+AvePool$^\dagger$ | $0.856 \pm 0.021$ | $0.935 \pm 0.010$ | $0.978 \pm 0.004$ | $0.793 \pm 0.044$ | $0.909 \pm 0.011$ | $0.951 \pm 0.005$ | $0.976 \pm 0.004$ | $0.984 \pm 0.004$ |
| NAN | $0.860 \pm 0.012$ | $0.933 \pm 0.009$ | **$0.979 \pm 0.004$** | $0.804 \pm 0.036$ | $0.909 \pm 0.013$ | $0.954 \pm 0.007$ | $0.978 \pm 0.004$ | $0.984 \pm 0.003$ |
| NAN$^\dagger$ | **$0.881 \pm 0.011$** | **$0.941 \pm 0.008$** | $0.978 \pm 0.003$ | **$0.817 \pm 0.041$** | **$0.917 \pm 0.009$** | **$0.958 \pm 0.005$** | **$0.980 \pm 0.005$** | **$0.986 \pm 0.003$** |

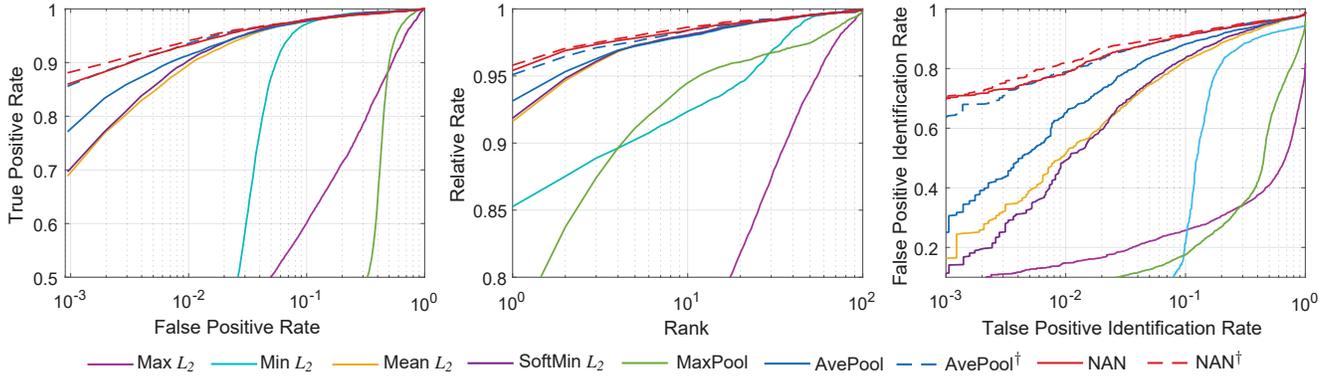

Figure 4. Average ROC (Left), CMC (Middle) and DET (Right) curves of the NAN and the baselines on the IJB-A dataset over 10 splits.

In this dataset, each training and testing instance is called a 'template', which comprises 1 to 190 mixed still images and video frames. Since one template may contain multiple medias and the dataset provides the media id for each image, another possible aggregation strategy is first aggregating the frame features in each media then the media features in the template [10, 30]. This strategy is also tested in this work with *CNN+AvePool* and our NAN. Note that media id may not be always available in practice.

We test the proposed method on both the 'compare' protocol for *1:1 face verification* and the 'search' protocol for *1:N face identification*. For verification, the true accept rates (TAR) vs. false positive rates (FAR) are reported. For identification, the true positive identification rate (TPIR) vs. false positive identification rate (FPIR) and the Rank-N accuracies are reported. Table 2 presents the numerical results of different methods, and Figure 4 shows the receiver operating characteristics (ROC) curves for verification as well as the cumulative match characteristic (CMC) and decision error trade-off (DET) curves for identification. The metrics are calculated according to [19, 13] on the 10 splits.

In general, the *CNN+Max $L_2$*, *CNN+Min $L_2$* and *CNN+MaxPool* perform worst among the baseline methods. *CNN+SoftMin $L_2$* performs slightly better than *CNN+MaxPool*. The use of media id significantly improves the performance of *CNN+AvePool*, but gives a relatively small boost to NAN. We believe NAN already has the robustness to templates dominated by poor images from a few media. Without the media aggregation, NAN outperforms all its baselines by appreciable margins, especially on the low FAR cases. For example, in the verification task, the TARs of our NAN at FARs of 0.001 and 0.01 are respectively 0.860 and 0.933, reducing the errors of the best results from its baselines by about 39% and 23%, respectively.

Table 3. Verification accuracy comparison of state-of-the-art methods, our baselines and NAN network on the YTF dataset.

| Method | Accuracy (%) | AUC |
|---|---|---|
| LM3L [16] | 81.3 ± 1.2 | 89.3 |
| DDML(combined)[15] | 82.3 ± 1.5 | 90.1 |
| EigenPEP [22] | 84.8 ± 1.4 | 92.6 |
| DeepFace-single [35] | 91.4 ± 1.1 | 96.3 |
| DeepID2+ [33] | 93.2 ± 0.2 | – |
| Wen et al. [41] | 94.9 | – |
| FaceNet [31] | 95.12 ± 0.39 | – |
| VGG-Face [28] | 97.3 | – |
| CNN+Max. $L_2$ | 91.96 ± 1.1 | 97.4 |
| CNN+Min. $L_2$ | 94.96 ± 0.79 | 98.5 |
| CNN+Mean $L_2$ | 95.30 ± 0.74 | 98.7 |
| CNN+SoftMin $L_2$ | 95.36 ± 0.77 | 98.7 |
| CNN+MaxPool | 88.36 ± 1.4 | 95.0 |
| CNN+AvePool | 95.20 ± 0.76 | 98.7 |
| NAN | **95.72 ± 0.64** | **98.8** |

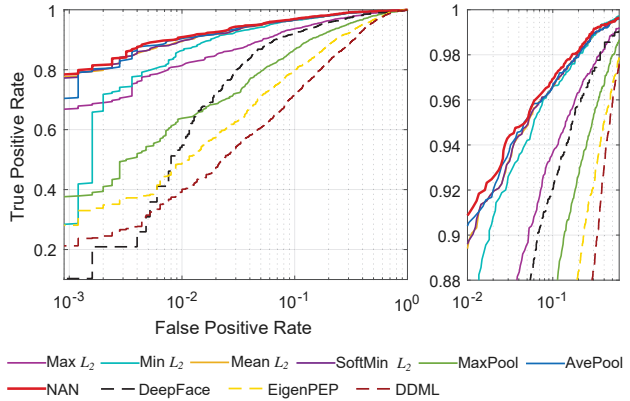

Figure 5. Average ROC curves of different methods and our NAN on the YTF dataset over the 10 splits.

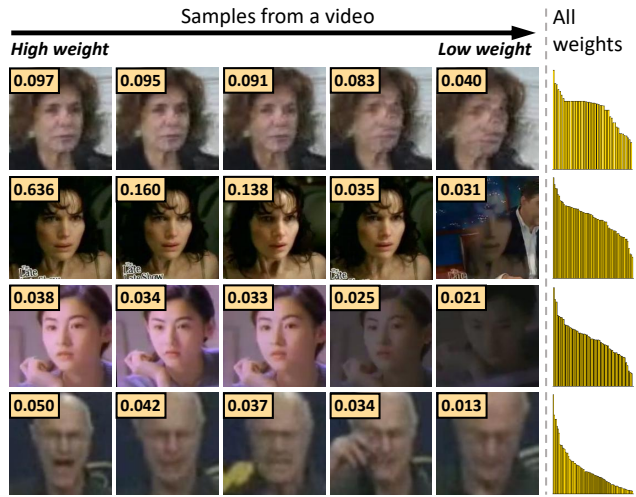

Figure 6. Typical examples on the YTF dataset showing the weights of the video frames computed by our NAN. In each row, five frames are sampled from a video and sorted based on their weights (numbers in the rectangles); the rightmost bar chart shows the sorted weights of all the frames (heights scaled).

To our knowledge, with the media aggregation our NAN achieves top performances compared to previous methods. It has a same verification TAR at FAR=0.1 and identification Rank-10 CMC as the state-of-the-art method of [10], but outperforms it on all other metrics (*e.g.* 0.881 *vs.* 0.836 TARs at FAR=0.01, 0.817 *vs.* 0.774 TPIRs at FPIR=0.01 and 0.958 *vs.* 0.928 Rank-1 accuracy).

Figure 3 has shown some typical examples of the weighting results. NAN exhibits the ability to choose high-quality and more discriminative face images while repelling poor face images.

### 3.4. Results on YouTube Face dataset

We then test our method on the YouTube Face (YTF) dataset [42] which is designed for unconstrained *face verification* in videos. It contains 3,425 videos of 1,595 different people, and the video lengths vary from 48 to 6,070 frames with an average length of 181.3 frames. Ten folds of 500 video pairs are available, and we follow the standard verification protocol to report the average accuracy with cross-validation. We again use the STN and similarity transformation to align the face images.

The results of our NAN, its baselines, and other methods are presented in Table 3, with their ROC curves shown in Fig. 5. It can be seen that the NAN again outperforms all its baselines. The gaps between NAN and the best-performing baselines are smaller compared to the results on IJB-A. This is because the face variations in this dataset are relatively small (compare the examples in Fig. 6 and Fig. 3), thus no much beneficial information can be extracted compared to naive average pooling or computing mean $L_2$ distances.

Compared to previous methods, our NAN achieves a mean accuracy of 95.72%, reducing the error of FaceNet by 12.3%. Note that FaceNet is also based on a GoogLeNet style network, and the average similarity of all pairs of 100 frames in each video (*i.e.*, 10K pairs) was used [31]. To our knowledge, only the VGG-Face [28] achieves an accuracy (97.3%) higher than ours. However, that result is based on a further discriminative metric learning on YTF, without which the accuracy is only 91.5% [28].

Table 4. Identification performance (rank-1 accuracies, %) on the Celebrity-1000 dataset for the *close-set* tests.

| Method | Number of Subjects | | | |
|---|---|---|---|---|
| | 100 | 200 | 500 | 1000 |
| MTJSR [23] | 50.60 | 40.80 | 35.46 | 30.04 |
| Eigen-PEP [22] | 50.60 | 45.02 | 39.97 | 31.94 |
| CNN+Mean $L_2$ | 85.26 | 77.59 | 74.57 | 67.91 |
| CNN+AvePool - VideoAggr | 86.06 | 82.38 | 80.48 | 74.26 |
| CNN+AvePool - SubjectAggr | 84.46 | 78.93 | 77.68 | 73.41 |
| NAN - VideoAggr | 88.04 | 82.95 | **82.27** | 76.24 |
| NAN - SubjectAggr | **90.44** | **83.33** | **82.27** | **77.17** |

### 3.5. Results on Celebrity-1000 dataset

The Celebrity-1000 dataset [23] is designed to study the unconstrained video-based *face identification* problem. It contains 159,726 video sequences of 1,000 human subjects, with 2.4M frames in total (∼15 frames per sequence). We use the provided 5 facial landmarks to align the face images. Two types of protocols – open-set and close-set - exist on this dataset. More details about the protocols and the dataset can be found in [23].

**Close-set tests.** For the close-set protocol, we first train the network on the video sequences with the identification loss. We take the FC layer output values as the scores and the subject with the maximum score as the result. We also train a linear classifier for *CNN+AvePool* to classify each video feature. As the features are built on video sequences, we call this approach 'VideoAggr' to distinguish it from another approach to be described next. Each subject in the dataset has multiple video sequences, thus we can build a single representation for a subject by aggregating all available images in all the training (gallery) video sequences. We call this approach 'SubjectAggr'. This way, the linear classifier can be bypassed, and identification can be achieved simply by comparing the feature $L_2$ distances.

The results are presented in Table 4. Note that [23] and [22] are not using deep learning and no deep network based method reported result on this dataset. So we mainly compare with our baselines in the following. It can be seen from Table 4 and Fig. 7 (a) that NAN consistently outperforms the baseline methods for both 'VideoAggr' and 'SubjectAggr'. Significant improvements upon the baseline are achieved for the 'SubjectAggr' approach. It is interesting to see that, 'SubjectAggr' leads to a clear performance drop for *CNN+AvePool* compared to its 'VideoAggr'. This indicates that the naive aggregation gets even worse when applied on the subject level with multiple videos. However, our NAN can benefit from 'SubjectAggr', yielding results consistently better than or on par with the 'VideoAggr' approach and delivers a considerable accuracy boost compared to the baseline. This suggests our NAN works quite well on handling large data variations.

Table 5. Identification performance (rank-1 accuracies, %) on the Celebrity-1000 dataset for the *open-set* tests.

| Method | Number of Subjects | | | |
|---|---|---|---|---|
| | 100 | 200 | 400 | 800 |
| MTJSR [23] | 46.12 | 39.84 | 37.51 | 33.50 |
| Eigen-PEP [22] | 51.55 | 46.15 | 42.33 | 35.90 |
| CNN+Mean $L_2$ | 84.88 | 79.88 | 76.76 | 70.67 |
| CNN+AvePool - SubjectAggr | 84.11 | 79.09 | 78.40 | 75.12 |
| NAN - SubjectAggr | **88.76** | **85.21** | **82.74** | **79.87** |

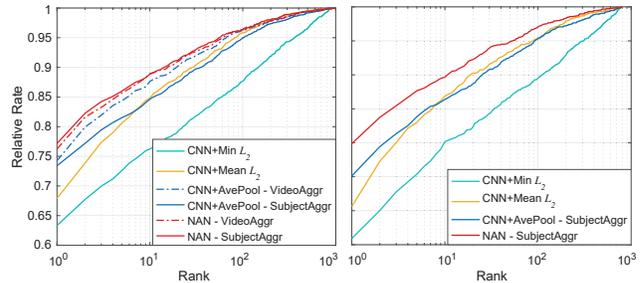

(a) Close-set tests on 1000 subjects  (b) Open-set tests on 800 subjects
Figure 7. The CMC curves of different methods on Celebrity 1000.

**Open-set tests.** We then test our NAN with the *close-set* protocol. We first train the network on the provided training video sequences. In the testing stage, we take the 'SubjectAggr' approach described before to build a highly-compact face representation for each gallery subject. Identification is perform simply by comparing the $L_2$ distances between aggregated face representations.

The results in both Table 5 and Fig. 7 (b) show that our NAN significantly reduces the error of the baseline *CNN+AvePool*. This again suggests that in the presence of large face variances, the widely used strategies such as average-pooling aggregation and the pairwise distance computation are far from optimal. In such cases, our learned NAN model is clearly more powerful, and the aggregated feature representation by it is more favorable for the video face recognition task.

### 4. Conclusions

We have presented a Neural Aggregation Network for video face representation and recognition. It fuses all input frames with a set of content adaptive weights, resulting in a compact representation that is invariant to the input frame order. The aggregation scheme is simple with small computation and memory footprints, but can generate quality face representations after training. The proposed NAN can be used for general video or set representation, and we plan to apply it to other vision tasks in our future work.

**Acknowledgments** GH was partly supported by NSFC Grant 61629301. HL's work was supported in part by Australia ARC Centre of Excellence for Robotic Vision (CE140100016) and by CSIRO Data61.